\documentclass[sigconf]{acmart}

\AtBeginDocument{%
  \providecommand\BibTeX{{%
    \normalfont B\kern-0.5em{\scshape i\kern-0.25em b}\kern-0.8em\TeX}}}
    
\newcommand{\mc}{\multicolumn}
    
\setcopyright{rightsretained}
\copyrightyear{2021}
\acmYear{2021}
\acmDOI{10.1145/1122445.1122456}

\acmConference[ICAIL 2021]{ICAIL 2021: International Conference on Artificial Intelligence and Law}{June 21--25, 2021}{São Paulo, Brazil}
\acmPrice{15.00}
\acmISBN{978-1-4503-XXXX-X/18/06}

\usepackage{color, colortbl}
\usepackage{hhline}
\usepackage{soul}
\definecolor{Green}{rgb}{0.8,1,0.8}
\definecolor{Red}{rgb}{1,0.8,0.8}
\definecolor{H1}{rgb}{0.796, 0.835, 0.91}
\definecolor{H2}{rgb}{0.99, 0.80, 0.67}
\definecolor{H3}{rgb}{0.7, 0.89, 0.8}
\newcolumntype{g}{>{\columncolor{H3}}r}
\DeclareRobustCommand{\hlblue}[1]{{\sethlcolor{H1}\hl{#1}}}
\DeclareRobustCommand{\hlorange}[1]{{\sethlcolor{H2}\hl{#1}}}
\DeclareRobustCommand{\hlgreen}[1]{{\sethlcolor{H3}\hl{#1}}}

\begin{document}

\title{Lex Rosetta: Transfer of Predictive Models Across Languages, Jurisdictions, and Legal Domains}

\author{Jaromir Savelka}
\email{jsavelka@cs.cmu.edu}
\affiliation{
  \institution{Carnegie Mellon University}
  \country{USA}
}

\author{Hannes Westermann}
\author{Karim Benyekhlef}
\affiliation{%
  \institution{Universit\'e de Montr\'eal}
  \country{Canada}}

\author{Charlotte S. Alexander}
\author{Jayla C. Grant}
\affiliation{%
 \institution{Georgia State University}
 \country{USA}}

\author{David Restrepo Amariles}
\author{Rajaa El Hamdani}
\affiliation{%
  \institution{HEC Paris}
  \country{France}
}

\author{S\'ebastien Mee\`us}
\author{Aurore Troussel}
\affiliation{%
  \institution{HEC Paris}
  \country{France}
}

\author{Micha\l\ Araszkiewicz}
\affiliation{%
  \institution{Uniwersytet Jagiello\'nski}
  \country{Poland}}

\author{Kevin D. Ashley}
\author{Alexandra Ashley}
\affiliation{%
  \institution{University of Pittsburgh}
  \country{USA}}

\author{Karl Branting}
\affiliation{%
  \institution{MITRE Corporation}
  \country{USA}}

\author{Mattia Falduti}
\affiliation{%
  \institution{Libera Universit\`a di Bolzano}
  \country{Italy}}
  
\author{Matthias Grabmair}
\affiliation{%
  \institution{Technische Universit\"at M\"unchen}
  \country{Germany}}
  
\author{Jakub Hara\v{s}ta}
\author{Tereza Novotn\'a}
\affiliation{%
  \institution{Masarykova univerzita}
  \country{Czech Republic}}
  
\author{Elizabeth Tippett}
\author{Shiwanni Johnson}
\affiliation{%
  \institution{University of Oregon}
  \country{USA}}

\renewcommand{\shortauthors}{Savelka and Westermann, et al.}

\begin{abstract}
 In this paper, we examine the use of multi-lingual sentence embeddings to transfer predictive models for functional segmentation of adjudicatory decisions across jurisdictions, legal systems (common and civil law), languages, and domains (i.e. contexts). 
 Mechanisms for utilizing  linguistic resources outside of their original context have significant potential benefits in AI \& Law because differences between legal systems, languages, or traditions often block wider adoption of research outcomes. 
We analyze the use of Language-Agnostic Sentence Representations in sequence labeling models using Gated Recurrent Units (GRUs) that are transferable across languages. To investigate transfer between different contexts we developed an annotation scheme for functional segmentation of adjudicatory decisions. We found that models generalize beyond the contexts on which they were trained (e.g., a model trained on administrative decisions from the US can be applied to criminal law decisions from Italy). 
Further, we found that training the models on multiple contexts increases robustness and improves overall performance when evaluating on previously unseen contexts. Finally, we found that pooling the training data from all the contexts enhances the models' in-context performance. 

\end{abstract}

\copyrightyear{2021}
\acmYear{2021}
\acmConference[ICAIL'21]{Eighteenth International Conference for Artificial Intelligence and Law}{June 21--25, 2021}{São Paulo, Brazil}
\acmBooktitle{Eighteenth International Conference for Artificial Intelligence and Law (ICAIL'21), June 21--25, 2021, São Paulo, Brazil}\acmDOI{10.1145/3462757.3466149}
\acmISBN{978-1-4503-8526-8/21/06}




\maketitle

\section{Introduction}
This paper explores the ability of multi-lingual sentence embeddings to enable training of predictive models that generalize beyond individual languages, legal systems, jurisdictions, and domains (i.e., contexts). We propose a new type schema for functional segmentation of adjudicatory decisions (i.e., decisions of trial and appellate court judges, arbitrators, administrative judges and boards) and use it to annotate legal cases across eight different contexts (7 countries, 6 languages). We release the newly created dataset (807 documents with 89,661 annotated sentences) including the annotation schema to the public.\footnote{\url{https://github.com/lexrosetta/caselaw_functional_segmentation_multilingual}}


In the area of AI \& Law,  research typically focuses on a single context, such as decisions of a specific court on a specific issue within a specific time range. This is justified by the complexity of legal work and the need for nuanced solutions to particular problems. At the same time, this narrow focus can limit the applicability of the research outcomes, since a proposed solution might not be readily transferable to a different context. In text classification, for example, a model might simply memorize a particular vocabulary characteristic of a given context, rather than acquiring the semantics of a predicted type. Adaptation of such a model to a new context would then require the assembly of a completely new dataset. This may be both time-consuming and expensive, since the annotation of legal documents relies on legal expertise.


Certain tasks appear to be of interest to researchers from multiple countries with different legal traditions (e.g., deontic classification of legal norms embodied in statutory law, argument extraction from case law, summarization/simplification of legal documents, etc.). This suggests that there may be several core tasks in AI \& Law that are of general interest in almost any context. One such task is a functional segmentation of adjudicatory decisions, which has been the subject of numerous studies in the past (see Section \ref{sec:related_work}). 
In this paper, we show that for this particular task it is possible to leverage linguistic resources created in multiple contexts.

This has wide-reaching implications for AI \& Law research. Since annotation of training data is expensive, models that are able to use existing data from other contexts might be instrumental in enabling real-world applications that can be applied across contexts. Such approaches may further enable international collaboration of researchers, each annotating their own part of a dataset to contribute to a common pool (as we do in this work) that could be used to train strong models able to generalize across contexts.



\subsection{Functional Segmentation}
We investigate the task of segmenting adjudicatory decisions based on the functional role played by their parts. While there are significant differences in how decisions are written in different contexts, we hypothesize that selected elements might be universal, such as, for example, sections:

\begin{enumerate}
    \item describing facts that give rise to a dispute;
    \item applying general legal rules to such facts;
    \item stating an outcome of the case (i.e., how was it decided).
\end{enumerate}

\noindent This conjecture is supported by the results of the comparative project titled Interpreting Precedents \cite{maccormick2016interpreting}, which aimed to analyze (among other things) the structure
in 11 different jurisdictions.
The findings of this project suggest that the structure indicated above may be considered a general model followed in the investigated jurisdictions, although variations exist that are characteristic of particular legal systems and types of courts and their decisions. 



The ability to segment cases automatically could be beneficial for many tasks. It could support reading and understanding of legal decisions by students, legal practitioners, researchers, and the public. It could facilitate empirical analyses of the discourse structure of decisions. It could enhance the performance, as well as the user experience, of legal search tools. For example, if a user searches for an application of a legal rule, they might restrict the search to the section where a judge applies the rule to a factual situation. Judges themselves might find the technique useful, within their own jurisdictions but also in transnational disputes involving the application of different legal standards. The same benefits apply to non-court settings, e.g., international arbitration, where many jurisdictions' laws and their interpretation matter. Further, the segmentation of decisions into meaningful sections could serve as an important step in many legal document processing pipelines.


\subsection{Hypotheses}
\label{sec:hypotheses}
To investigate how well predictive models, based on multi-lingual sentence embeddings, learn to segment cases into functional parts across different contexts, we evaluated the following hypotheses:
\begin{enumerate}
    \item[(H1)] A model trained on a single context can generalize when transferred to other, previously unseen, contexts.
    \item[(H2)] A model trained on data pooled from multiple contexts is more robust and generalizes better to unseen contexts than a model trained on a single context.
    \item[(H3)] A context-specific model benefits from pooling the in-domain data with data from other contexts.
\end{enumerate}

\subsection{Contributions}
\label{sec:contributions}
By carrying out this work, we provide the following contributions to the AI \& Law research community:
\begin{itemize}
    \item Detailed definition and analysis of a functional segmentation task that is widely applicable across different contexts.
    \item A new labeled dataset consisting of 807 documents (89,661 sentences) from seven countries in six different languages.
    \item Evidence of the effectiveness of multi-lingual embeddings on processing legal documents.
    \item Release of the code used for data preparation, analysis, and the experiments in this work.
\end{itemize}

\section{Related Work}
\label{sec:related_work}
Segmenting court decisions into smaller elements according to their function or role is an important task in legal text processing. Prior research utilizing supervised machine learning (ML) approaches or expert crafted rules can roughly be distinguished into two categories. First, the task could be to segment the text into a small number of contiguous parts typically comprising multiple paragraphs (this work). Different variations of this task were applied to several legal domains from countries, such as Canada~\cite{farzindar2004letsum}, the Czech Republic \cite{harasta2019automatic}, France~\cite{boniolperformance}, or the U.S. \cite{savelka2018segmenting}. Second, the task could instead be labeling smaller textual units, often sentences, according to some predefined type system (e.g., rhetorical roles, such as evidence, reasoning, conclusion).  Examples from several domains and countries include administrative decisions from the U.S.~\cite{walker2019automatic,zhong2019automatic}, multi-domain court decisions from India \cite{bhattacharya2019identification}, international arbitration decisions \cite{branting2019semi}, or even multi-\{domain,country\} adjudicatory decisions in English \cite{savelka2020cross}. Identifying a section that states an outcome of the case has also received considerable attention separately \cite{xu2020using,petrova2020extracting}. To the best of our knowledge, existing work on functional segmentation of court decisions is limited to a single language---ours being the first paper exploring the task jointly on legal documents in multiple languages.

In NLP, the success of word embeddings was followed by an increasing interest in learning continuous vector representations of longer linguistic units, such as sentences (a trend that has been reflected in AI~\&~Law research as well \cite{zhong2019automatic,westermann2020paragraph}). 
Multi-lingual representations recently attracted ample attention. 
While most of the earlier work was limited to a few close languages or pairwise joint embeddings for English and one foreign language, several approaches to obtain general-purpose massively multi-lingual sentence representations were proposed \cite{artetxe2019massively,devlin2019bert,conneau2020unsupervised}. Such representations were utilized in many downstream applications, such as document classification~\cite{lai2019bridging}, machine translation \cite{aharoni2019massively}, question answering~\cite{lewis2020mlqa},
hate speech detection \cite{aluru2020deep}, or information retrieval (IR) in the legal domain \cite{zhebel2020different}. Our work is one of the first such applications in the legal domain and to the best of our knowledge the first dealing with more than two languages.


Approaches other than language-agnostic sentence embeddings (this work) were used in AI \& Law research focused on texts in multiple languages. A recent line of work mapped recitals to articles in EU directives and normative provisions in Member states' legislation \cite{nanda2020multilingual}. There, mono-lingual models were used (i.e., one model per language). Other published applications in multi-lingual legal IR were based on thesauri \cite{dini2005cross,sheridan1997cross}. A common technique to bridge the language gap was the use of ontologies and knowledge graphs \cite{boella2015linking,agnoloni2007building,gonzalez2018business,ajani2016european}. The multi-lingual environments, such as EU or systems established by international treaties, attracted work on machine translation \cite{koehn2009machine}, meaning equivalence verification \cite{tang2019verifying}, and building of parallel corpora \cite{steinberger2014overview,sugisaki2016building}.



\section{Dataset}

\begin{table*}[t]
  \setlength{\tabcolsep}{1.7pt}
  \footnotesize
  \caption{Descriptive statistics of the created dataset. Each entry provides information about the country, the language of the decisions (Lang), and the number of documents (Docs) in a specific context. The Sentence-Level Statistics subsection reports basic descriptive statistics focused on sentences as well as the number of sentences labeled with each type (OoS - Out of Scope, Head - Heading, Int.S. - Introductory Summary, Back - Background, Anl - Analysis, Out - Outcome). The part highlighted in green contains the counts of sentences labeled with the types we focus on in this work.}
  \label{tab:data_set}
  \begin{tabular}{|llr|rrrrrrr|ggg|l|}
    \hhline{~~~|----------|~}
    \mc{3}{c|}{} & \mc{10}{c|}{\cellcolor{gray!10}Sentence-Level Statistics} \\
    \hhline{|---|~~~~~~~~~~|-|}
    \rowcolor{gray!10}Country &Lang &Docs &Count & Avg & Min & Max   &OoS    &Head &\mc{1}{l}{Int.S.} &Back &Anl  &\mc{1}{l|}{Out}  & Description \\
    \hline
    Canada   &EN  &100  &12168 &121.7&8    &888  &873 &438  &20     &3319  &7190 &328  & Random selection of cases retrieved from \url{www.canlii.org} from multiple provinces.      \\
             &    &     &      &     &     &     &{\tiny 7.2\%}&{\tiny 3.6\%}&{\tiny 0.2\%}&{\tiny 27.3\%}&{\tiny 59.1\%}&{\tiny 2.7\%}& The selection is not limited to any specific topic or court.                                     \\
    \hline
    Czech R. &CS  &100  &11283 &112.8&10   &701  &945 &1257 &2      &3379 &5422 &278  & A random selection of cases from Constitutional Court (30), Supreme Court (40), and             \\
             &    &     &      &     &     &     &{\tiny 8.4\%}&{\tiny 11.1\%}&{\tiny 0.0\%}&{\tiny 29.9\%}&{\tiny 48.1\%}&{\tiny 2.5\%}& Supreme Administrative Court (30). Temporal distribution was taken into account.                \\
    \hline
    France   &FR  &100  &5507  &55.1 &8    &583  &3811  &220  &0      &485   &631  &360  & A selection of cases decided by Cour de cassation between 2011 and 2019. A stratified           \\
             &    &     &      &     &     &     &{\tiny 69.2\%}&{\tiny 4.0\%}&{\tiny 0.0\%}&{\tiny 8.8\%}&{\tiny 11.4\%}&{\tiny 6.5\%}& sampling based on the year of publication of the decision was used to select the cases.         \\
    \hline
    Germany  &DE  &104  &10724 &103.1&12   &806 &406 &333  &38     &2960 &6697 &290  & A stratified sample from the federal jurisprudence database spanning all federal courts         \\
             &    &     &      &     &     &     &{\tiny 3.8\%}&{\tiny 3.1\%}&{\tiny 0.4\%}&{\tiny 27.6\%}&{\tiny 62.4\%}&{\tiny 2.7\%}& (civil, criminal, labor, finance, patent, social, constitutional, and administrative).          \\
    \hline
    Italy    &IT  &100  &4534  &45.3 &10   &207  &417 &1098 &0      &986  &1903 &130  & The top 100 cases of the criminal courts stored between 2015 and 2020 mentioning                \\
             &    &     &      &     &     &     &{\tiny 9.2\%}&{\tiny 24.2\%}&{\tiny 0.0\%}&{\tiny 21.7\%}&{\tiny 42.0\%}&{\tiny 2.9\%}& ``stalking'' and keyed to the Article 612 bis of the Criminal Code.                             \\
    \hline
    Poland   &PL  &101  &9791 &96.9&4    &1232 &796 &303  &0      &2736 &5820 &136  & A stratified sample from trial-level, appellate, administrative courts,  the Supreme Court,          \\
             &    &     &      &     &     &     &{\tiny 8.1\%}&{\tiny 3.1\%}&{\tiny 0\%}      &{\tiny 27.9\%}&{\tiny 59.4\%}&{\tiny 1.4\%}&  and the Constitutional tribunal. The cases mention ``democratic country ruled by law.''                 \\
    \hline
    U.S.A. I &EN  &102  &24898 &244.1&34   &1121 &574&1235 &475     &6042 &16098&474  & Federal district court decisions in employment law mentioning ``motion for summary                \\
             &    &     &      &     &     &     &{\tiny 2.3\%}&{\tiny 5.0\%}&{\tiny 1.9\%}&{\tiny 24.3\%}&{\tiny 64.7\%}&{\tiny 1.9\%}& judgment," ``employee,'' and ``independent contractor.''                                        \\
    \hline
    U.S.A. II&EN  &100  &10756 &107.6&24   &397  &1766&650  &639 &3075 &4402 &224  & Administrative decisions from the U.S. Department of Labor. Top 100 ordered in        \\
             &    &     &      &     &     &     &{\tiny 1.6\%}&{\tiny 6.0\%}&{\tiny 5.9\%}&{\tiny 28.6\%}&{\tiny 40.9\%}&{\tiny 2.1\%}&  reverse chronological rulings order, starting in October 2020, were selected.                \\
    \hline
    Overall  &6   &807  &89661&105.6&4    &1232 &9588&5534 &1174    &22982 &48163&2220 &\mc{1}{c}{}             \\
    \cline{1-13}
  \end{tabular}
\end{table*}

In creating the dataset, the first goal was to identify a task that would be useful across different contexts. After extensive literature review, we identified the task of functional segmentation of adjudicatory decisions as a viable candidate. 
To make the task generalizable, we decided to include only a small number of core types.

\begin{enumerate}
    \item \emph{Out of Scope} -- Parts outside of the main document body (e.g., metadata, editorial content, dissents, end notes, appendices).
    \item \emph{Heading} -- Typically an incomplete sentence or marker starting a section (e.g., ``Discussion,'' ``Analysis,'' ``II.'').
    \item \emph{Background} -- The part where the court describes procedural history, relevant facts, or the parties' claims.
    \item \emph{Analysis} -- The section containing reasoning of the court, issues, and application of law to the facts of the case.
    \item \emph{Introductory Summary} -- A brief summary of the case at the beginning of the decision.
    \item \emph{Outcome} -- A few sentences stating how the case was decided (i.e, the overall outcome of the case).
\end{enumerate}

We created detailed annotation guidelines defining the individual types as well as describing the annotation workflow (tooling, steps taken during annotation). Eight teams of researchers from six different countries (14 persons) were trained in the annotation process through online meetings. After this, each annotator conducted a dry-run annotation on 10 cases and received detailed feedback. Then, each team was tasked with assembling approximately 100 adjudicatory decisions. Each team developed specifications for the decisions to be included in their part of the dataset. 

Four of the contexts were double-annotated by two annotators (Canada, Czech R., France, U.S.A. I); the remaining four by just one. Each team had at least one member with a completed law degree. When a team had more than one member, law students were allowed to be included. 


A high-level description of the resulting dataset is provided in Table \ref{tab:data_set}. It consists of eight contexts from seven different countries (two parts are from the U.S.) with 807 documents in six languages (three parts are in English). Most of the contexts include judicial decisions, while U.S.A. II was the only context that consisted solely of administrative decisions.
There are considerable variations in the length of the documents. While an average document in the U.S.A.~I context comprises of 530.6 sentences, an average document in the France context is about ten times shorter (59.0 sentences).

The four double-annotated parts enabled us to examine the inter-annotator agreement. Table \ref{tab:ia_agreement_2} shows the raw agreement on a character level. 
While it appears that recognizing the \emph{Outcome} was rather straightforward in the France and U.S.A. I contexts, it was more complicated in case of Canada and the Czech R. This might be due to a presence/absence of some structural clue. We also observe that in the Czech R. context it was presumably much easier to distinguish between the \emph{Background} and \emph{Analysis} than in case of the other three contexts.

\begin{table}[t]
    \footnotesize
    \caption{Raw agreement on a character level for the four datasets with two human annotators. The agreement is computed as a percentage of characters where both the annotators agree on a specific type over all the characters annotated by that type by any of the annotators. (NM=Not Marked)}
    \label{tab:ia_agreement_2}
    \centering
    \begin{tabular}{lrrrrrrr}
    \toprule
             & OoS & Head & Int.S. & Back & Anl  & Out & NM \\
    \midrule
    Canada   &97.2 & 68.2 & 44.0   & 83.3 & 92.2 & 79.9 & 43.4    \\
    Czech R. &80.3 & 54.6 & 0.0    & 92.6 & 94.5 & 46.9 & 10.0    \\
    France   &93.5 & 92.5 & N/A    & 43.0 & 72.2 & 99.1 & 1.0    \\
    U.S.A. I &90.8 & 71.0 & 74.2   & 78.4 & 93.7 & 91.1 & 18.4    \\
    \midrule
    Overall  &91.8 & 70.4 & 72.1   & 82.1 & 92.6 & 77.3 & 3.8 \\
    \bottomrule
    \end{tabular}
\end{table}

In this paper, we focus on prediction of the \emph{Background}, \emph{Analysis}, and \emph{Outcome} types. We decided to exclude the \emph{Introductory Summary} type, since it is mainly present in the data from the United States. For the double-annotated datasets, we picked the annotations that appeared to be of higher quality (either by consensus between the annotators themselves or by a decision of a third unbiased expert).


We first removed all the spans of text annotated with either of the \emph{Out of Scope} or \emph{Heading} types. The removal of \emph{Out of Scope} leaves the main body of a decision stripped of potential metadata or editorial content at the beginning of the document as well as dissents, or end notes at the end. The removal of the text spans annotated with the \emph{Heading} type might appear counter-intuitive since headings often provide a clear clue as to the content of the following section (e.g., ``Outcome'', ``Analysis'' etc.).
We remove these (potentially valuable) headings because we want to focus on the more interesting task of recognizing the sections purely by the semantics of their constitutive sentences. This task is more challenging, and more closely emulates generalization to domains where headings are not used or not present in all cases, or are not reliable indicators.



The transformed documents are separated into several segments based on the annotations of the three remaining types. Each segment is then split into sentences.\footnote{We used the processing pipeline from \url{https://spacy.io/} (large models). For the Czech language we used \url{https://github.com/TakeLab/spacy-udpipe} with the Czech model (PDT) from \url{https://universaldependencies.org/}. The output was further processed with several regular expressions. A different method was used for the French dataset, which consists of a few very long sections, internally separated by a semicolon. After consultation with an expert we decided to split the cases by the semicolon as well.} A resulting document is a sequence of sentences labeled with one of the \emph{Background}, \emph{Analysis}, or \emph{Outcome} types. The highlighted (green) part of Table \ref{tab:data_set} provides basic descriptive statistics of the resulting dataset per the individual contexts. Our final dataset for analysis consists of 807 cases split into 74,539 annotated sentences.


\section{Models}
\label{sec:models}
In our experiments we use the Language-Agnostic Sentence Representations (LASER) model \cite{artetxe2019massively} to encode sentences from different languages into a shared semantic space. Each document becomes a series of vectors which represent the semantic content of a single sentence. We use these vectors to train a bidirectional Gated Recurrent Unit (GRU) model~\cite{cho2014learning} for predicting sentence labels.

The LASER model is a language-agnostic bidirectional LSTM encoder coupled with an auxiliary decoder and trained on parallel corpora. The sentence embeddings are obtained by applying a max-pooling operation over the output of the encoder. The resulting sentence representations (after concatenating both directions) are 1024-dimensional. The released trained model,\footnote{\url{https://github.com/facebookresearch/LASER}} which we use in this work, supports 93 languages (including the six in our dataset)  belonging to 30 different families and  written in 28 different scripts. The model was trained on 223 million parallel sentences. The joint encoder itself has no information on the language or writing script of the tokenized text, while the tokenizer is language specific. It is even possible to mix multiple languages in one sentence. 
The focus of the LASER model is to produce vector representations of sentences that are general with respect to two dimensions: the input language and the NLP task \cite{artetxe2019massively}. An interesting property of such universal multi-lingual sentence embeddings is the increased focus on the sentence semantics, as the syntax or other surface properties are unlikely to be shared among languages.

The GRU neural network \cite{cho2014learning} is an architecture based on a recurrent neural network (RNN) that is able to learn the mapping from a sequence of an arbitrary length to another sequence. GRUs are able to either score a pair of sequences or to generate a target sequence given a source sequence (this work). In a bidirectional GRU, two separate sequences are considered (one from right to left and the other from left to right). Traditional RNNs work well for shorter sequences but cannot be successfully applied to long sequences due to the well-known problem of vanishing gradients. Long Short-Term Memory (LSTM) networks \cite{hochreiter1997long} have been used as an effective solution to this problem (the forget gate, along with the additive property of the cell state gradients). GRUs have been proposed as an alternative to LSTMs with a reduced number of parameters. In GRUs there is no explicit memory unit, and the forget gate and the update gate are combined. The performance of GRUs was shown to be superior to that of LSTMs in the scenario of long texts and small datasets \cite{yang2020lstm}, which is the situation in this work. For these reasons, we chose to use GRUs over LSTMs.

\begin{figure}[t] 
    \centering
    \includegraphics[width=7cm]{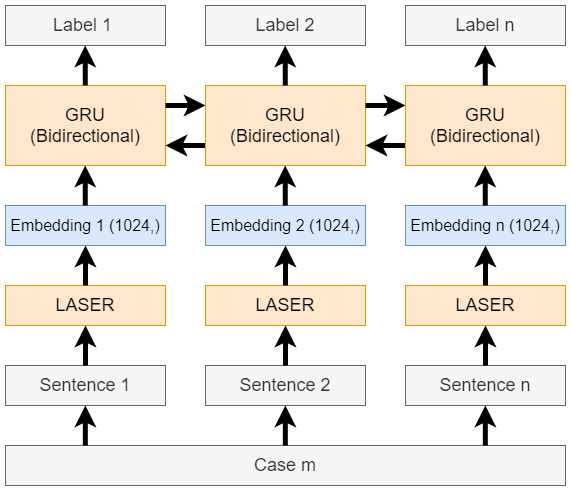}
    \caption{The structure of the sequential model used for prediction. Each case $m$ is split into $n$ sentences, which are converted to language-independent LASER vector embeddings. These are fed through a bidirectional recurrent GRU-model, which predicts one of the three labels per input sentence.}
    \label{fig:model}
\end{figure}

The overall structure of the employed model is shown in Figure~\ref{fig:model}.\footnote{The model was implemented using the Keras framework (\url{https://keras.io/}).} Each case is transformed into a $1080\times1024$ matrix. The number 1080 represents the maximum length (in sentences) of any case in our dataset. Shorter cases are padded to be of uniform length. The vectors are passed to the model in batches of size 32. They first go through a masking layer, which masks the sentences used for padding. The data is then passed to a bidirectional GRU model with 256 units and a dropout of $0.2$. Finally, the model contains a time-distributed dense output layer with softmax activation, which outputs the predicted label (i.e., \emph{Background}, \emph{Analysis}, \emph{Outcome}) for each sentence. As a loss function, we use categorical cross-entropy. As optimizer we use the Adam algorithm \cite{kingma2014adam} with initial learning rate set to $4e^{-3}$ (reduced to $4e^{-4}$ once validation loss has stopped decreasing, with a patience of 50 epochs). We train the model for up to 1000 epochs. We halt the training once the validation accuracy has not increased in 80 epochs. For prediction, we use the best model as determined by validation accuracy.

\section{Experimental Design}
\label{sec:design}
We performed three experiments to test the hypotheses (Section~\ref{sec:hypotheses}). The first experiment (H1) focused on model generalization across different contexts (Out-Context). The second experiment~(H2) assessed model robustness when trained on multiple pooled contexts different from the one where the model was applied (Pooled Out-Context). Finally, the third experiment (H3) analyzed the effects of pooling the target context data with data from other contexts (Pooled with In-Context). The different pools of training data, and the baselines we compare them against, are summarized in Table \ref{tab:model_configs} and described in Sections \ref{sec:exp_out_context}-\ref{sec:exp_pooled_w_in_context}.

Since our dataset is limited, we performed a 10-fold cross-vali\-dation. The folds were kept consistent across all the experiments. The experiments were conducted in the following manner:

\begin{enumerate}
    \item Index $i=1$ is set.
    \item Pool of training context(s) is selected (see Sections \ref{sec:exp_out_context}-\ref{sec:exp_pooled_w_in_context}).
    \item A single test context is selected (see Sections \ref{sec:exp_out_context}-\ref{sec:exp_pooled_w_in_context}).
    \item Eight of the folds from the training context(s) are used as training data (index different from $i$ and $(i+1) \bmod 10$).
    \item The folds with index $(i+1) \bmod 10$ from the training context(s) are designated as validation data.
    \item The $i$ fold from the test context is designated as test data.
    \item The models are trained and evaluated.
    \item Index $i=i+1$ is set.
    \item If $i\leq10$ go to (4) else finish.
\end{enumerate}

\noindent Note that from here on we highlight the baselines using the colors as shown in Table \ref{tab:model_configs} to improve clarity.


\begin{figure}[t] 
    \centering
    \includegraphics[width=8.5cm]{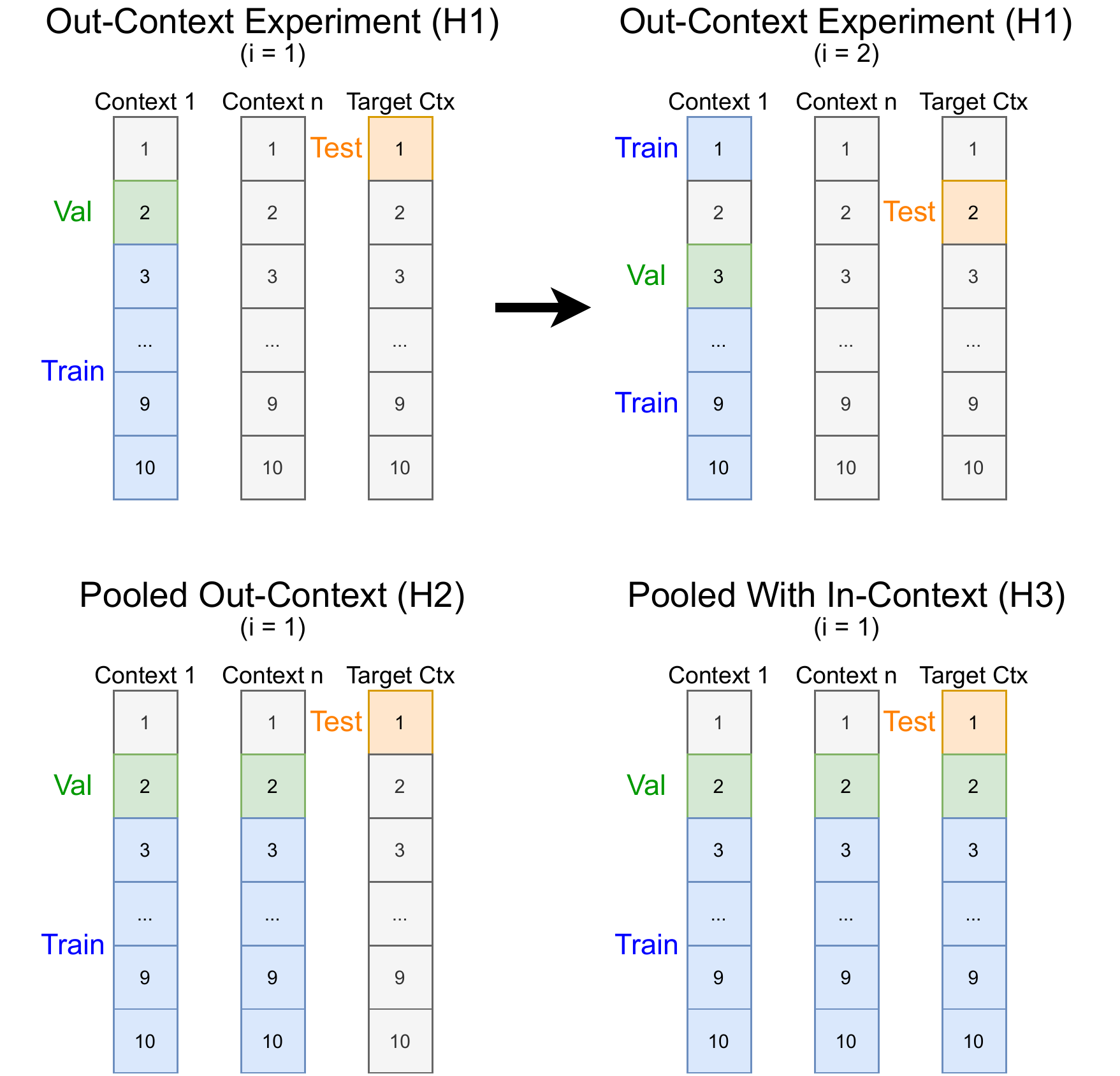}
    \caption{A visualization of the three experimental setups. The figure shows how training contexts are selected for H1, H2 and H3, and examples of how the folds are assigned to Train, Val and Test.}
    \label{fig:exp_design}
\end{figure}

\begin{table}[t]
    \footnotesize
    \setlength{\tabcolsep}{1.7pt}
    \caption{Description of different training data selection and baselines used for H1, H2 and H3. The \hlblue{random} (H1), the \hlorange{best single Out-Context models} (H2), and the \hlgreen{In-Context models} (H3) baselines are highlighted.}
    \label{tab:model_configs}
    \centering
    \begin{tabular}{llll}
    \toprule
    Name         & Trained on & Hyp. & Baseline \\
    \midrule
    \cellcolor{H1}Random   & Target Context& - & -    \\
    \cellcolor{H3}In-Context & Target Context & - & - \\
    Out-Context & Non-Target Context & H1 & \cellcolor{H1}Random \\
    Pooled Out-Context & Pooled Non-Target & H2 & \cellcolor{H2}Best performing Out-\\
                       & Contexts         & & \cellcolor{H2}Context model per context\\
    Pooled with In-Context & Non-Target and Target & H3 &  \cellcolor{H3}In-Context \\
                           & Pooled Contexts\\
    \bottomrule
    \end{tabular}
\end{table}

\subsection{Out-Context Experiment (H1)}
\label{sec:exp_out_context}
In this experiment, we investigated the ability of the models to generalize beyond the training context. The pool of training data consisted of a single context at a time (see Figure \ref{fig:exp_design}).
As a baseline, we used a \hlblue{random} classifier that learns the distribution of the target dataset (train and validation folds) and generates predictions based on that distribution. The baseline thus had a slight advantage in having access to the signal from the target dataset. The Out-Context models performing better than the random baseline would show that the models are able to transfer (at least some) knowledge from one context to another. 


\subsection{Pooled Out-Context Experiment (H2)}
\label{sec:exp_pooled}
The second experiment focused on the ability of the models to gain robustness as they learn from more than one context, when applied to an unseen context. Therefore, the training pool for the experiments consisted of data from all the contexts \emph{except} the current target context (see Figure \ref{fig:exp_design}). 
As a baseline for each context, we used the performance measurement taken from the \hlorange{best Out-Context model for that context} (see Section \ref{sec:exp_out_context}). This was a very competitive baseline since the best Out-Context model likely stems from the context that is the most similar to the target context. If the Pooled Out-Context model performed better than the best Out-Context model, this would indicate that pooling datasets increases the robustness of the model, allowing it to perform better on previously unseen contexts. 

\subsection{Pooled With In-Context Experiment (H3)}
\label{sec:exp_pooled_w_in_context}
The third experiment focused on pooling the target context's data with data from the other contexts. The training pool therefore incorporates all of the contexts, including the target one (see Figure~\ref{fig:exp_design}).
As a baseline, we used the \hlgreen{In-Context models}, trained on the target context. Again, this is a very strong baseline since the In-Context model should be able to learn the most accurate signal from the target context (including context specific peculiarities). If the model trained on the pooled contexts (including target) is able to outperform the In-Context model, this would indicate that pooling contexts is beneficial in terms of both robustness and absolute performance on the individual contexts.



\section{Evaluation Method}
\label{sec:evaluation}
To evaluate the performance of the systems trained on data from different contexts, we used Precision ($P$), Recall ($R$), and $F_1$-measure.


\noindent 
We compute the evaluation metrics for each class (i.e., \emph{Background}, \emph{Analysis}, \emph{Outcome}) per fold. We then report per class $F_1$-scores as well as an overall $F_1$-score (micro) including its standard deviation.


For statistical evaluation we used the Wilcoxon signed-rank test~\cite{wilcoxon1992individual} to perform pair comparisons between methods and the baselines as suggested by \cite{demsar2006statistical}. We used the overall (micro) $F_1$\nobreakdash-score as the evaluation metric. 
The null-hypothesis states that the mean performance of the two compared methods (i.e., an assessed model and a baseline) is equivalent.

Since the number of samples is rather small (7 contexts in testing H1, and 8 contexts in testing H2 and H3), we are not able to reject the null hypotheses at $\alpha=0.05$ for any of H1--H3. This is as expected given the size of the dataset, and it is not unacceptable given the exploratory nature of our work at this stage. We will be in a position to formally evaluate the hypotheses once we extend the dataset (see Section \ref{sec:future_work}). Instead, we report raw p-values produced by the testing as an evidence of the assessed methods' effectiveness. The p-value is the probability of seeing the results at least as extreme as those we observed given the null hypothesis is correct (i.e., their mean performance is the same). 

\section{Results}
\label{sec:results}

\begin{table*}[]
    \centering
    \caption{Results of the Out-Context, Pooled Out-Context, and Pooled with In-Context experiments. Each row reports a performance of a model across contexts. A bold number in each cell reports the (micro) average F$_1$-score over the predicted classes and the standard deviation across the 10 folds. The F$_1$-scores of the three classes are reported below ordered as Background, Analysis, Outcome. A visual explanation of cell contents can be found in Figure \ref{fig:table_expl}. The \hlblue{random} (H1), the \hlorange{best single Out-Context models} (H2), and the \hlgreen{In-Context} (H3) baselines are highlighted. The p-values are color coded to match the baselines.}
    
    \setlength{\tabcolsep}{4.5pt}
    \label{tab:results}
    \begin{tabular}{lllllllllll}
    \toprule
       &\mc{1}{c}{Canada}&\mc{1}{c}{Czech R.}&\mc{1}{c}{France}&\mc{1}{c}{Germany}& \mc{1}{c}{Italy} & \mc{1}{c}{Poland}& \mc{1}{c}{U.S.A. I} & \mc{1}{c}{U.S.A. II} & \mc{1}{c}{Avg (-test)} & \mc{1}{c}{Avg (+test)}\\
    \midrule
Random	&\cellcolor{H1}$\mathbf{.54\pm.04}$&\cellcolor{H1}$\mathbf{.49\pm.02}$&\cellcolor{H1}$\mathbf{.33\pm.05}$&\cellcolor{H1}$\mathbf{.56\pm.04}$&\cellcolor{H1}$\mathbf{.50\pm.04}$&\cellcolor{H1}$\mathbf{.55\pm.04}$&\cellcolor{H1}$\mathbf{.58\pm.03}$&\cellcolor{H1}$\mathbf{.51\pm.02}$&\mc{2}{c}{$\mathbf{.51\pm.07}$}	\\
(dist.)	&.31 .66 .06 &.36 .59 .04 &.32 .37 .25 &.29 .68 .03 &.31 .62 .04 &.32 .67 .00 &.26 .72 .02 &.37 .61 .02 &\mc{2}{c}{.32 .62 .10}\\
\midrule
Canada&\cellcolor{H3}$\mathbf{.82\pm.09}$&	$\mathbf{.68\pm.08}$&	$\cellcolor{H2}\mathbf{.64\pm.09}$&	$\mathbf{.81\pm.06}$&	$\mathbf{.73\pm.08}$&	$\mathbf{.73\pm.07}$&	$\cellcolor{H2}\mathbf{.87\pm.05}$&	$\mathbf{.88\pm.03}$&$\mathbf{.76\pm.09}$&$\mathbf{.77\pm.08}$	\\
\textcolor{H1}{$p=.016$}&.75 .87 .70 &.53 .80 .39 &.64 .66 .57 &.71 .89 .00 &.55 .82 .66 &.50 .85 .00 &.75 .92 .69 &.87 .90 .70 & .65 .83 .43 & .66 .84 .46 \\
\midrule
Czech R.&$\mathbf{.76\pm.09}$&	\cellcolor{H3}$\mathbf{.91\pm.04}$&	$\mathbf{.47\pm.08}$&	$\mathbf{.82\pm.08}$&	$\mathbf{.82\pm.05}$&	$\cellcolor{H2}\mathbf{.83\pm.07}$&	$\mathbf{.84\pm.06}$&	$\mathbf{.86\pm.05}$&$\mathbf{.77\pm.13}$&$\mathbf{.79\pm.13}$	\\
\textcolor{H1}{$p=.016$}&.71 .80 .31 &.90 .92 .64 &.52 .36 .48 &.75 .89 .01 &.81 .84 .40 &.73 .90 .00 &.73 .90 .28 &.86 .88 .41 & .73 .80 .27 & .75 .81 .32 \\
\midrule
France	&$\mathbf{.71\pm.07}$&	$\mathbf{.61\pm.06}$&	\cellcolor{H3}$\mathbf{.86\pm.08}$&	$\mathbf{.66\pm.10}$&	$\mathbf{.73\pm.08}$&	$\mathbf{.65\pm.07}$&	$\mathbf{.72\pm.07}$&	$\mathbf{.68\pm.08}$&$\mathbf{.68\pm.04}$&$\mathbf{.70\pm.07}$	\\
\textcolor{H1}{$p=.016$}&.45 .83 .69 &.37 .78 .45 &.81 .83 .98 &.37 .82 .00 &.59 .81 .69 &.33 .82 .00 &.32 .86 .69 &.48 .81 .73 & .42 .82 .46 & .47 .82 .53 \\
\midrule
Germany&$\mathbf{.72\pm.10}$&	$\mathbf{.64\pm.09}$&	$\mathbf{.29\pm.12}$&	\cellcolor{H3}$\mathbf{.88\pm.11}$&	$\mathbf{.69\pm.09}$&	$\mathbf{.77\pm.09}$&	$\mathbf{.73\pm.10}$&	$\mathbf{.83\pm.07}$&$\mathbf{.67\pm.16}$&$\mathbf{.69\pm.17}$	\\
\textcolor{H1}{$p=.031$}&.68 .76 .01 &.42 .81 .01 &.42 .32 .00 &.82 .93 .66 &.50 .84 .00 &.54 .88 .54 &.47 .85 .00 &.82 .88 .00 & .55 .76 .08 & .58 .78 .15 \\
\midrule
Italy&$\mathbf{.55\pm.12}$&	$\mathbf{.76\pm.09}$&	$\mathbf{.63\pm.08}$&	$\mathbf{.78\pm.09}$&	\cellcolor{H3}$\mathbf{.95\pm.02}$&	$\mathbf{.73\pm.10}$&	$\mathbf{.53\pm.13}$&	$\mathbf{.74\pm.08}$&$\mathbf{.67\pm.10}$&$\mathbf{.71\pm.13}$	\\
\textcolor{H1}{$p=.047$}&.57 .55 .49 &.74 .81 .12 &.69 .63 .52 &.73 .83 .00 &.92 .96 .94 &.66 .78 .00 &.50 .54 .42 &.74 .74 .63 & .66 .70 .31 & .69 .73 .39 \\
\midrule
Poland&$\mathbf{.76\pm.08}$&	$\cellcolor{H2}\mathbf{.83\pm.05}$&	$\mathbf{.38\pm.11}$&\cellcolor{H2}$\mathbf{.85\pm.08}$&	$\cellcolor{H2}\mathbf{.83\pm.05}$&	\cellcolor{H3}$\mathbf{.93\pm.05}$&	$\mathbf{.73\pm.08}$&	$\mathbf{.83\pm.07}$&$\mathbf{.74\pm.15}$&$\mathbf{.77\pm.16}$	\\
\textcolor{H1}{$p=.016$}&.66 .84 .00 &.82 .89 .01 &.48 .52 .00 &.73 .91 .44 &.80 .91 .00 &.89 .95 .88 &.44 .86 .00 &.82 .87 .00 & .68 .83 .06 & .71 .84 .17 \\
\midrule
U.S.A. I&\cellcolor{H2}$\mathbf{.83\pm.06}$&	$\mathbf{.65\pm.08}$&	$\mathbf{.47\pm.14}$&	$\mathbf{.81\pm.07}$&	$\mathbf{.65\pm.15}$&	$\mathbf{.67\pm.09}$&	\cellcolor{H3}$\mathbf{.91\pm.03}$&	$\cellcolor{H2}\mathbf{.89\pm.03}$&$\mathbf{.71\pm.13}$&$\mathbf{.74\pm.14}$	\\
\textcolor{H1}{$p=.016$}&.76 .87 .59 &.45 .79 .49 &.35 .61 .35 &.71 .89 .00 &.40 .80 .58 &.38 .83 .00 &.84 .94 .73 &.87 .91 .68 & .56 .81 .38 & .60 .83 .43 \\
\midrule
U.S.A. II&$\mathbf{.81\pm.08}$&	$\mathbf{.67\pm.06}$&	$\mathbf{.53\pm.15}$&	$\mathbf{.84\pm.10}$&	$\mathbf{.75\pm.10}$&	$\mathbf{.70\pm.08}$&	$\mathbf{.86\pm.05}$&	\cellcolor{H3}$\mathbf{.94\pm.02}$&$\mathbf{.74\pm.11}$&$\mathbf{.76\pm.12}$	\\
\textcolor{H1}{$p=.016$}&.74 .86 .65 &.49 .80 .31 &.50 .63 .41 &.76 .91 .00 &.57 .84 .75 &.43 .84 .00 &.72 .92 .59 &.93 .96 .82 & .60 .83 .39 & .64 .85 .44 \\
\midrule
Pooled &$\mathbf{.83\pm.06}$&	$\mathbf{.87\pm.03}$&	$\mathbf{.66\pm.08}$&	$\mathbf{.90\pm.04}$&	$\mathbf{.85\pm.04}$&	$\mathbf{.88\pm.05}$&	$\mathbf{.81\pm.10}$&	$\mathbf{.92\pm.03}$&\mc{2}{c}{$\mathbf{.84\pm.08}$}	\\
\textcolor{H2}{$p=.148$}&.77 .86 .66 &.87 .91 .03 &.59 .67 .71 &.87 .95 .01 &.81 .89 .65 &.83 .94 .01 &.64 .88 .73 &.91 .93 .65 &\mc{2}{c}{.79 .88 .43}\\
\midrule
Pooled+	&$\mathbf{.88\pm.05}$&	$\mathbf{.94\pm.03}$&	$\mathbf{.82\pm.09}$&	$\mathbf{.96\pm.02}$&	$\mathbf{.94\pm.04}$&	$\mathbf{.94\pm.04}$&	$\mathbf{.92\pm.03}$&	$\mathbf{.96\pm.02}$&\mc{2}{c}{$\mathbf{.92\pm.04}$}	\\
\textcolor{H3}{$p=.195$}&.83 .91 .77 &.94 .95 .70 &.76 .78 .96 &.94 .98 .64 &.91 .95 .90 &.92 .97 .65 &.86 .95 .84 &.95 .96 .80 &\mc{2}{c}{.89 .93 .78}\\

    \bottomrule
    \end{tabular}
\end{table*}

\begin{figure}[t] 
    \centering
    \includegraphics[width=.45\textwidth]{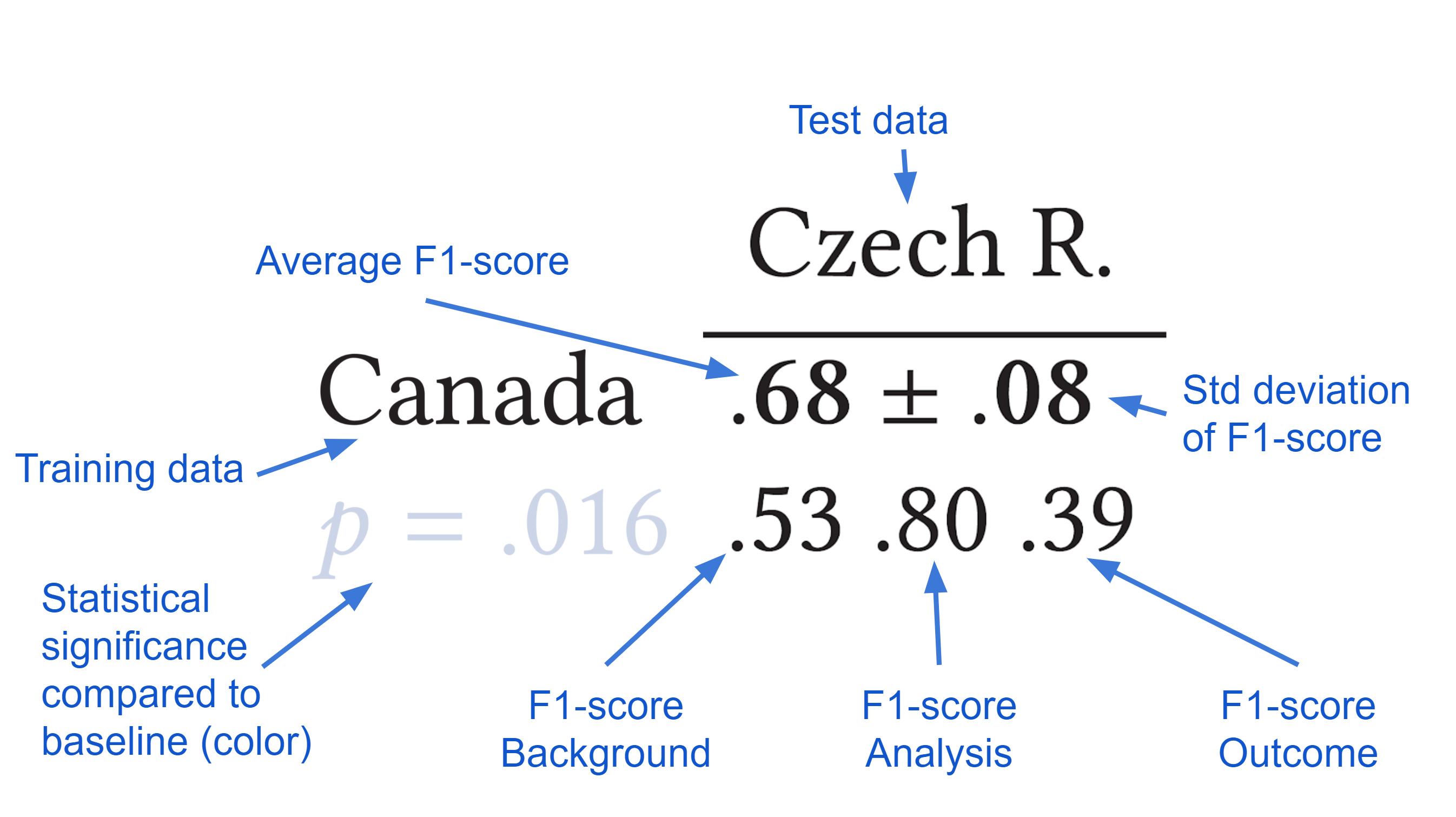}
    \caption{A description of performance metrics reported in Table \ref{tab:results}.}
    \label{fig:table_expl}
\end{figure}

The results of all three experiments are shown in Table \ref{tab:results}. Each row reports a performance of a specific model across different contexts (columns). Each cell shows a bold number on the first row which correspond to the (micro) average F$_1$-score over the three predicted classes, with standard deviation across the 10 folds. The F$_1$-scores of the three classes are reported in the second row of each cell ordered as \emph{Background}, \emph{Analysis}, \emph{Outcome}. A visual explanation of cell contents can be found in Figure \ref{fig:table_expl}.

\subsection{Out-Context Experiment (H1)}
The performance of the models trained during the Out-Context experiment is reported by the eight rows of Table \ref{tab:results} starting with Canada and ending with U.S.A. II. Here, the models are trained on a single context (row), and then evaluated on all of the contexts (columns). 
It appears the models perform reasonably well under the Out-Context condition. The application of the trained models outperforms the \hlblue{random baseline} across the board (in 54 out of 56 instances). This is further corroborated by the low p-values obtained for these methods when compared to the random baseline (reported in Table \ref{tab:results}).

Several interesting patterns emerge from the results. First, it appears that models trained on contexts with the same language or a language from the same family perform better. For example, the models trained on Canada and the two U.S.A. contexts perform well among each other. A similar observation applies to the models trained on the Poland and the Czech R. contexts. Quite surprisingly the models succeeded in identifying the \emph{Outcome} sentences to some extent, despite a heavy under-representation of these sentences in our dataset. For example, a model trained on Canada predicts the \emph{Outcome} sentences of the both U.S.A. contexts with an average F$_1$-score close to 0.7. At the same time the \emph{Outcome} sentences are by far the most challenging ones. On multiple occasions the models completely fail to predict these (e.g., Canada$\rightarrow$Germany, or Poland$\rightarrow$France).

Overall, the results demonstrate the ability of the models to effectively transfer the knowledge learned on one context to another. At the same time, it is clear that the models trained on one context and applied to a different one do not perform as well as the \hlgreen{In-Context models}.

\subsection{Pooled Out-Context Experiment (H2)}
The performance of Pooled Out-Context models is reported in the Pooled row of Table \ref{tab:results}. The experiment concerns the resulting models' robustness, i.e., if a model trained on multiple contexts adapts well to unseen contexts. We are especially interested if such a model adapts better than the models trained on single contexts.

The results suggest that training on multiple contexts leads to models that are robust and perform better than the models trained on a single context. The multi-context models outperform the \hlorange{best single extra-context model baseline} in 7 out of 8 cases. The $p=0.148$ needs to be understood in terms of the small number of samples (contexts) and competitiveness of the baseline. 

Interestingly, the Pooled Out-Context models even appear to be competitive with several \hlgreen{In-Context models} (Canada, Czech R., Germany, U.S.A. II). The overall average F$_1$-scores are often quite high (over 0.80 or 0.90). This is a surprising outcome considering the fact that no data from the context on which a model is evaluated is used during training.

\subsection{Pooled With In-Context Experiment (H3)}
The performance of the Pooled with In-Context model is reported in the Pooled+ row of Table \ref{tab:results}. This experiment models a scenario where a sample of labeled data from the target context is available. The question is whether combining In-Context data with data from other contexts leads to improved performance.

The results appear to suggest that pooling the target context with data from other contexts does lead to improved performance. In case of 3 out of the 8 contexts (Canada, Czech R., and U.S.A. I) the improvement is clear and substantial across all the three classes (\emph{Background}, \emph{Analysis}, \emph{Outcome}). For three additional contexts (Germany, Poland, and U.S.A. II), the performance also improved in terms of overall (micro) F$_1$-score, but took a slight hit for the challenging \emph{Outcome} class. With respect to  the  two remaining contexts (France, Italy) the overall performance of the pooled models is lower than that of the \hlgreen{In-Context models}. As in the previous experiment, the $p=0.195$ needs to be understood in terms of the small number of samples (contexts) and high competitiveness of the In-Context baseline. 



\section{Discussion}
\label{sec:discussion}
It appears that the multilingual sentence embeddings generated by the LASER model excel at capturing the semantics of a sentence. A model trained on a single context could in theory capture the specific vocabulary used in that context. This would almost certainly lead to poor generalization across different contexts. However, the performance statistics we observed when transferring the Out-Context models to other contexts suggest that the sentence embeddings provide a representation of the sentences that enable the model to learn aspects of the meaning of a sentence, rather than mere surface linguistic features.

The results clearly point to certain relationships where contexts within the same or related languages appear to work well together, e.g., \{Canada, U.S.A. I, and U.S.A. II\} or \{Czech R. and Poland\}. This could indicate that the multi-lingual embeddings work better when the language is the same or similar. It could also point to similarities in legal traditions, e.g. the use of somewhat similar sentences to indicate a transition from one type of section to the next. Note that we removed headings, which means that explicit cues could not be relied on by the models we trained on such transformed documents. Finally, the cause could also be topical (domain) similarity of the contexts (e.g., both the U.S.A. contexts deal with employment law). Also note that the above are just possible explanations. We did not perform feature importance analysis on the models.

\begin{figure}[t] 
    \centering
    \includegraphics[width=.4\textwidth]{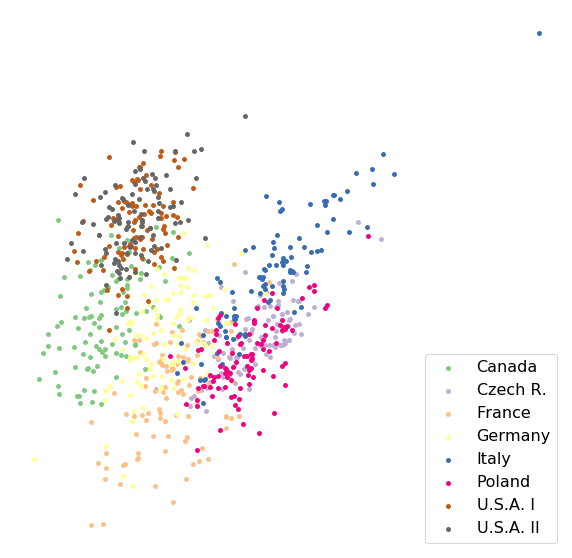}
    \caption{Average LASER embeddings for each case document, projected to 2-dimensional space using Principal Component Analysis.}
    \label{fig:avg_emb}
\end{figure}

To gain insight into this phenomenon, we visualized the relationships among the contexts on a document level. We first calculated the average sentence embedding for each document. This yielded 1024-dimensional vectors for 807 documents representing their semantics. We arranged the resulting vectors in a matrix ($1024\times807$) and performed a Principal Component Analysis (PCA) reducing the dimensionality of the document vectors to 2. This operation enabled a convenient visualization shown in Figure \ref{fig:avg_emb}.

Overall, the cases from the same contexts appear to cluster together. This is expected as the documents written in the same language, having similar topics, or sharing some similarities due to legal traditions, are likely to map to vectors that are closer. The documents from  both the U.S.A. contexts occupy the same region in the embedding space. This is not surprising as they come from the same jurisdiction, are written in the same language, and deal with similar topics (employment law). The Canadian cases, which are also in English, occupy a nearby space. This could be linked to the language as well as likely similarities in legal traditions. French, German, and Italian documents occupy the middle space; they are closer to the English documents than those from the Czech R. and Poland. Interestingly, the Czech and Polish documents occupy almost the same space. The Polish context focuses on the rule of law while the Czech one is supposed to be more general. As the latter deals with the decisions of the top-tier courts (one of them Constitutional), it is possible that the topics substantially overlap. Moreover, Poland and the Czech R. share similar legal traditions and languages from the same family (Slavic), so the close proximity of the documents in the embedding space might not be unexpected. Finally, German, French, and Canadian cases occupy wider areas than documents from other contexts. This could be due to their lack of focus on specific legal domains.

\begin{figure}[t] 
    \centering
    \includegraphics[width=.45\textwidth]{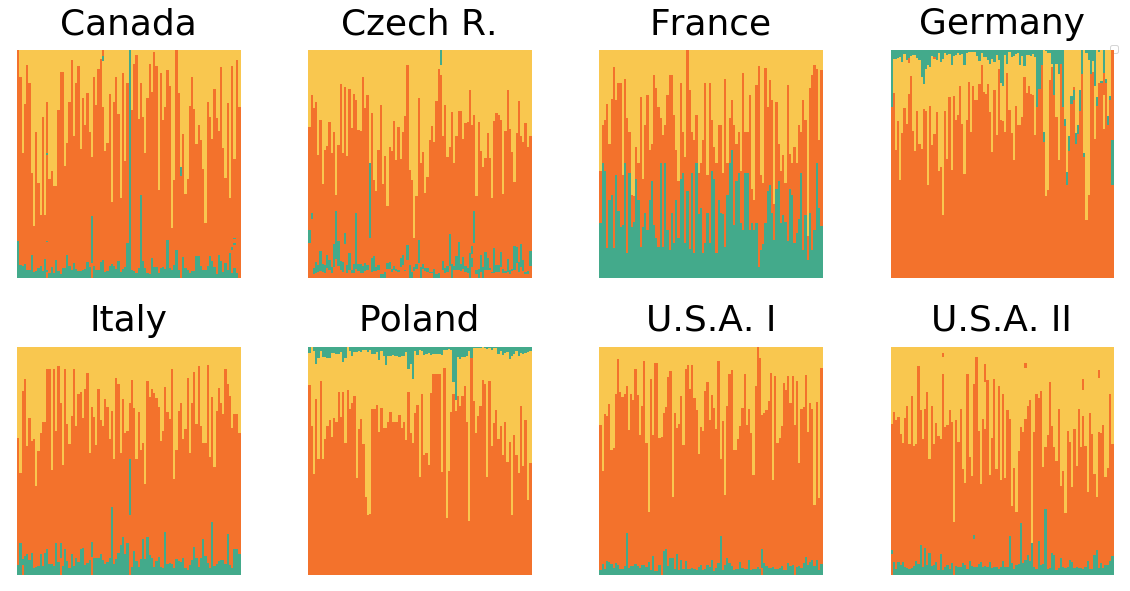}
    \caption{Distribution of labels across datasets. The X-axis corresponds to  unique cases, whereas the Y-axis corresponds to the normalized length of the cases. The colors correspond to the different labels of the sentences (Background, Analysis, Outcome).}
    \label{fig:label_dist}
\end{figure}

We observed a peculiar phenomenon where the Out-Context models trained on the German and Polish contexts failed to detect \emph{Outcome} sentences on the six remaining contexts and vice-versa. The cause is readily apparent from the visualization shown in Figure \ref{fig:label_dist}. Each segment of the figure depicts a spatial distribution of sentences color coded with their labels across documents for a particular context. As can be seen, the cases typically follow a pattern of a contiguous \emph{Background} segment, followed by a long \emph{Analysis} section. The several \emph{Outcome} sentences are placed at the very end of the documents. In the Polish and German decisions, however,  the \emph{Outcome} sentences come first. The GRU models we use rely on the structure as well as semantics in making their predictions. As we can see, a model trained exclusively on cases that begin with a \emph{Background} might therefore have difficulties correctly identifying outcome sections at the beginning, and vice-versa. However, as we will see below, a model trained with data featuring both structures can learn to correctly identify the correct structure based on the semantics of the sentences.


The model trained on the French context appears to perform better on detecting \emph{Outcome} sentences than models trained on other contexts. This is somewhat surprising as the French model's overall performance is among the weakest (e.g. compare the Czech model's average F$_1=0.77$ to the F$_1=0.68$ of the French model). Again, Figure \ref{fig:label_dist} provides an insight into why  this happens. The French context is the only one where the count of \emph{Outcome} sentences is comparable to those of the other two categories. For all the other contexts, the \emph{Outcome} sentences are heavily underrepresented. This reveals an interesting direction for future work where the use of re-sampling may yield models with better sensitivity for identifying  \emph{Outcome} sentences.

In two instances models trained on a single context under-per\-formed the \hlblue{random baseline}. The model trained on the Germany context achieved the average F$_1=0.29$ when applied to the French context (Random $F_1=0.33$). As the model trained on Polish data also performed poorly on the French context ($F_1=0.38$) the cause appears to be the inability of the two models to detect the \emph{Outcome} sentences at the end (discussed above). As the \emph{Outcome} sentences are heavily present in the French context, this problem manifests in a score lower than the random baseline. The second instance is the model trained on the Italy context applied to the U.S.A. I data (F$_1=0.53$ versus F$_1=0.58$ on Random). Here, the cause appears to be different. Note that the Italian context likely has a very specific notion of \emph{Outcome} sentences (F$_1=0.94$ on Italy$\rightarrow$Italy). It appears that many \emph{Analysis} sentences from the U.S.A. I context were labeled as \emph{Outcome} by the model. Summary judgments often address multiple legal issues with their own conclusions which could have triggered the model to label such sentences as \emph{Outcome}.

An important finding is the performance of the Pooled Out-Context model (H2) shown in the Pooled row of Table \ref{tab:results}. The experiment simulates training of a model on several contexts, and then applying it to an unseen context. The Pooled Out-Context models, having no access to the data from a target context, reliably outperform the \hlorange{best single Out-Context models}. They appear to be competitive with several \hlgreen{In-Context models}. These results are achieved with a fairly small dataset of 807 cases. We expect that expanding the dataset would lead to further improved performance.


The Pooled with In-Context experiment (H3) models the situation where data from a target context is available in addition to labeled data from other contexts. Our experiments indicate that the use of data from other contexts (if available) in addition to data from the target context is preferable to the use of the data from the target context only. This is evidenced by the improved performance of the models trained on the pooled contexts over the single \hlgreen{In-Context models}. The models have an interesting property of being able to identify the \emph{Outcome} sentences with effectiveness comparable to (or higher than) the models trained on the same context only. This holds for all the contexts, except Poland where the \emph{Outcome} performance is a bit lower (0.65 vs. 0.88). This indicates that the model is able to learn the two possible modes of the \emph{Outcome} section placement. It successfully distinguishes cases where the section is at the beginning from the cases where the \emph{Outcome} sentences are found toward the end.

The inclusion of the In-Context data in the pooled data leads to a remarkable improvement over only using the pooled Out-Context data. The magnitude of the improvement highlights the importance of including such data in the training. We envision that the models trained on different contexts used in combination with high-speed similarity annotation frameworks \cite{westermann2019computer,westermann2020sentence} could enable highly cost efficient annotation in situations where resources are scarce. Perhaps, adapting a model to an entirely new context could be as simple as starting with a model trained on other contexts, and spending a few hours correcting the misconceptions of the model to teach it the particularities of the new context.  

\section{Conclusions}
We analyzed the use of multi-lingual sentence embeddings in sequence labeling models to enable transfer across languages, jurisdictions, legal systems (common and civil law), and domains. We created a new type schema for functional segmentation of adjudicatory decisions and used it to annotate legal cases across eight different contexts. We found that models generalize beyond the contexts they were trained on and that training the models on multiple contexts increases their robustness and improves the overall performance when evaluating on previously unseen contexts. We also found that pooling the training data of a model with data from additional contexts enhances its performance on the target context. The results are promising in enabling re-use of annotated data across contexts and creating generalizable and robust models. We release the newly created dataset (807 documents with 89,661 annotated sentences), including the annotation schema and the code used in our experiments, to the public.

This work suggests a promising path for the future of international collaboration in the field of AI \& Law. While previous annotation efforts have typically been limited to a single context, the experiments presented here suggest that researchers can work together by annotating cases from many different contexts at the same time. Such a combined effort could aid researchers in creating models that perform well on the data from the context they care about, while at the same time helping other groups train even better models for other contexts. We encourage these research directions and hope to form such collaborations under the Lex Rosetta project.

\section{Future Work}
\label{sec:future_work}
The application of multi-lingual sentence embeddings to functional segmentation of case law across different contexts yielded promising results. At the same time, the work is subject to limitations and leaves much room for improvement. Hence, we suggest several directions for future work:

\begin{itemize}
    \item Extension of the datasets from different contexts used in this work beyond \textasciitilde{}100 documents per context. 
    \item Annotation of data from contexts beyond the eight used here (multi-lingual models support close to 100 languages).
    \item Analysis of automatic detection of \emph{Introductory Summary}, \emph{Headings}, and \emph{Out of Scope}.
    \item Identification and investigation of other tasks applicable across different contexts. 
    \item Evaluation of the application of other multilingual models (e.g., those mentioned in Section \ref{sec:related_work}).
    \item Exploring other transfer learning strategies beyond simple data pooling, such as the framework proposed in \cite{savelka2015transfer}.
    \item Using multi-lingual models for annotation tasks with high-speed annotation framework, such as \cite{westermann2019computer,westermann2020sentence}.
    \item Performing the transfer across contexts with related (but different) tasks, such as in \cite{savelka2020cross}.
    \item Further exploring the differences in the distribution of the multilingual embeddings for purposes of comparing and analyzing domains, languages, and legal traditions.
\end{itemize}

\begin{acks}
Hannes Westermann, Karim Benyeklef, and Kevin D. Ashley would like to thank the Cyberjustice Laboratory at Université de Montréal, the LexUM Chair on Legal Information and the Autonomy through Cyberjustice Technologies (ACT) project for their support of this research. Kevin D. Ashley also thanks the Canadian Legal Information Institute for providing the corpus of legal cases. Matthias Grabmair thanks the SINC GmbH for supporting this research. Jakub Hara\v{s}ta and Tereza Novotn\'{a} acknowledge the support of the ERDF project ``Internal grant agency of Masaryk University'' (No.~CZ.02.2.69/0.0/0.0/19\_073/0016943). 
\end{acks}

\bibliographystyle{ACM-Reference-Format}
\bibliography{sample-base}

\end{document}